\documentclass[lettersize,journal]{IEEEtran}
\usepackage{amsmath,amsfonts}
\usepackage{algorithmic}
\usepackage{algorithm}
\usepackage{array}
\usepackage[caption=false,font=normalsize,labelfont=sf,textfont=sf]{subfig}
\usepackage{textcomp}
\usepackage{stfloats}
\usepackage{url}
\usepackage{verbatim}
\usepackage{graphicx}
\usepackage{cite}

\usepackage{multirow}

\begin{document}

\title{InitialGAN: A Language GAN with Completely Random Initialization}

\author{Da Ren, Qing Li \\
The Hong Kong Polytechnic University
\thanks{Corresponding to Qing Li (csqli@comp.polyu.edu.hk)}}

\markboth{Journal of \LaTeX\ Class Files,~Vol.~14, No.~8, August~2021}%
{Shell \MakeLowercase{\textit{et al.}}: A Sample Article Using IEEEtran.cls for IEEE Journals}


\maketitle

\begin{abstract}
  Text generative models trained via Maximum Likelihood Estimation (MLE) suffer from the notorious exposure bias problem, and Generative Adversarial Networks (GANs) are shown to have potential to tackle this problem. Existing language GANs adopt estimators like REINFORCE or continuous relaxations to model word probabilities. The inherent limitations of such estimators lead current models to rely on pre-training techniques (MLE pre-training or pre-trained embeddings). Representation modeling methods which are free from those limitations, however, are seldomly explored because of their poor performance in previous attempts. Our analyses reveal that invalid sampling methods and unhealthy gradients are the main contributors to such unsatisfactory performance. In this work, we present two techniques to tackle these problems: dropout sampling and fully normalized LSTM. Based on these two techniques, we propose InitialGAN whose parameters are randomly initialized in full. Besides, we introduce a new evaluation metric, Least Coverage Rate, to better evaluate the quality of generated samples. The experimental results demonstrate that InitialGAN outperforms both MLE and other compared models. To the best of our knowledge, it is the first time a language GAN can outperform MLE without using any pre-training techniques.
\end{abstract}


\section{Introduction}

Text generative models are stepping stones for various natural language processing tasks~\cite{DBLP:conf/aaai/ZhangTY21, DBLP:conf/emnlp/MaL21}. Implementing Maximum Likelihood Estimation (MLE) with autoregressive structure has gained great success~\cite{DBLP:conf/nips/SutskeverVL14, DBLP:conf/icml/GehringAGYD17, DBLP:conf/nips/VaswaniSPUJGKP17}. This method uses ground truth as input during training, but reads previously generated tokens during inference. The discrepancy between training and inference, however, causes the exposure bias problem~\cite{DBLP:conf/nips/BengioVJS15, DBLP:conf/nips/GoyalLZZCB16, DBLP:conf/acl/ZhangFMYL19}. This problem affects the quality of generated sentences and grows the needs of exploring other alternatives in text generation. Generative Adversarial Networks (GANs)~\cite{DBLP:journals/corr/GoodfellowPMXWOCB14} are central in many image generation success stories~\cite{DBLP:conf/iclr/BrockDS19, DBLP:conf/cvpr/KarrasLAHLA20, DBLP:journals/corr/abs-2106-12423}. GANs can tackle the exposure bias problem by providing a consistent generation manner in training and inference.

However, the non-differentiable sampling operations in text generators stop gradients from passing through to generators, which limit the direct applications of GANs in text generation~\cite{DBLP:conf/aaai/YuZWY17}. Currently, many researchers tackle this problem by REINFORCE~\cite{DBLP:journals/ml/Williams92} or continuous relaxations~
\cite{DBLP:conf/iclr/MaddisonMT17, DBLP:conf/iclr/JangGP17}. REINFORCE is an unbias but high variance estimator~\cite{DBLP:journals/corr/abs-2011-13527}, whereas continuous relaxations are low variance but bias estimators~\cite{DBLP:conf/nips/dAutumeMRR19}. The inherent limitations of these two methods lead the fragile training of GANs to be more unstable, so existing language GANs rely on either MLE pre-training or pre-trained embedding comparable with MLE~\cite{ DBLP:journals/corr/abs-2011-13527, DBLP:conf/nips/dAutumeMRR19, DBLP:conf/nips/ScialomDLPS20}.

Methods based on REINFORCE or continuous relaxations explicitly model word probabilities at each timestep, so we denote them as \textbf{Probability Modeling Methods (PMMs)}. Another type of methods is to firstly transform words into representations, and then train generators to model these representations. We denote these methods as \textbf{Representation Modeling Methods (RMMs)}. Research on RMMs is extremely limited, due to the unsatisfactory performance in previous attempts~\cite{DBLP:conf/nips/dAutumeMRR19, kumar2020endtoend}. However, such methods should be a promising research line, since they contain neither non-differentiable operations nor bias estimators. The discrepancy between theoretical feasibility and unexpected poor performance prompts us to conduct an in-depth analysis of its reasons, based on which two problems are found as responsible for the poor performance of RMMs.

The first one is called ``invalid sampling'' problem. RMMs do not have word probabilities that can be sampled. Injecting random noise into generators is also demonstrated as ineffective in autoregressive structure~\cite{DBLP:conf/iclr/RazaviOPV19, DBLP:conf/acl/ZhuBLMLW20}. Generators with an invalid sampling method will generate samples in high similarities, and leads to the mode collapse problem~\cite{DBLP:conf/icml/ArjovskyCB17,DBLP:conf/nips/HeuselRUNH17}. Another problem is unhealthy gradients. RMMs update generators based on gradients from discriminators. Compared with other sequence models~\cite{DBLP:conf/naacl/PetersNIGCLZ18}, more layers are stacked to build the discriminator and the generator, so RMMs place higher demands on healthy gradients. Gradient vanishment is more severe in LSTM~\cite{DBLP:journals/neco/HochreiterS97}, for the output gate there further narrows down the gradients from other layers. Unhealthy gradients will directly influence the performance of generators.

To tackle the first problem, we introduce a simple but effective sampling method: \textbf{dropout sampling}. Unlike injecting random noise, it provides a non-negligible random factor by masking a certain number of dimensions in input. This method improves both diversity and quality of generated samples via relieving the mode collapse problem.

We solve the second problem by proposing a new variant of LSTM: \textbf{fully normalized LSTM}. Our theoretical analyses show that incorporating layer normalization~\cite{DBLP:journals/corr/BaKH16} in the calculation of hidden state can relieve gradient vanishment by providing an additional augmentation term in its derivative. This operation, however, is omitted in the existing combination of layer normalization and LSTM~\cite{DBLP:journals/corr/BaKH16}. Fully normalized LSTM makes up this shortcoming by simultaneously obtaining strong sequence modeling capabilities and healthier gradients.

Theoretically, language GANs can get satisfactory performance without any pre-training techniques (MLE pre-training or pre-trained embedding). In this article, we present \textbf{InitialGAN} to echo this significant goal in text generation. The contributions of this work can be summarized as follows:

\begin{itemize}
  \item We provide in-depth analysis and offer affective solutions to the two main limitations of representation modeling methods. For the invalid sampling problem, we introduce dropout sampling which is a simple but effective sampling method to improve both the quality and diversity of generated samples. For the unhealthy gradient problem, we propose a fully normalized LSTM which can relieve gradient vanishment by making use of layer normalization to provide an augmentation term.

  \item We put forward InitialGAN as a representation modeling based language GAN which is characterized by having all the parameters to be initialized randomly. In particular, InitalGAN has three models: aligner, generator and discriminator. The aligner transforms words into representations, and the generator tries to model word representations; the discriminator uses these representations as input and identify whether the representations are from the aligner or the generator. Different from existing language GANs which are based on pre-training techniques, all the parameters in InitialGAN are initialized randomly.

  \item Observing that the existing embedding level metric is not sensitive to the change of sample quality, we propose a new metric: Least Coverage Rate, which can better identify the differences among different models. The experimental results show that InitialGAN outperforms both MLE and other compared models. To the best of our knowledge, it is the first time a language GAN can outperform MLE without using any pre-training techniques. It also demonstrates that RMMs denote a promising research line for language GANs.
\end{itemize}

\section{Related Work}

Most of existing text generative models implement Maximum Likelihood Estimation (MLE) by combining cross entropy and an autoregressive structure~\cite{DBLP:conf/nips/SutskeverVL14, DBLP:conf/icml/GehringAGYD17, DBLP:conf/nips/VaswaniSPUJGKP17}. However, this method uses true tokens as input during training, and predicts based on previously generated tokens during inference. Mistakes from previous generation will lead the model to be in the state space that it has never seen during training, so they will be amplified quickly and lead to a sharp decrease of sentence quality~\cite{DBLP:conf/nips/BengioVJS15}. This problem is denoted as exposure bias~\cite{DBLP:conf/nips/GoyalLZZCB16, DBLP:conf/acl/AroraABC22}.

\begin{figure}[htbp]
  \centerline{\includegraphics[scale=0.24]{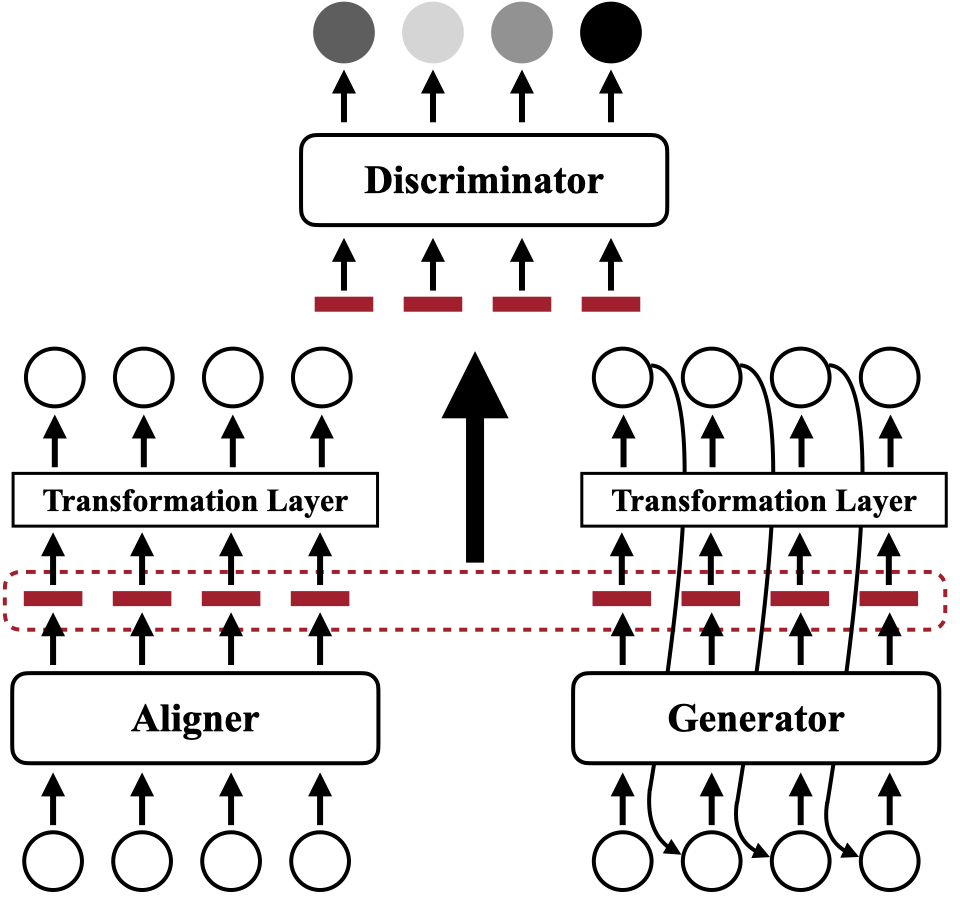}}
  \caption{Structure of InitialGAN.}
  \label{fig:structure}
\end{figure}

To tackle this problem, researchers try to incorporate GANs~\cite{DBLP:journals/corr/GoodfellowPMXWOCB14} into text generation. GANs have gained great success in various applications~\cite{DBLP:journals/tnn/TangLXTS23,DBLP:journals/tnn/MaoYW23,DBLP:journals/tnn/YeoSPK22,DBLP:journals/tnn/KimPH22,DBLP:journals/corr/abs-2301-09515,DBLP:journals/tnn/Zhang23}. However, its development in text generation is relatively slow. The main difficulty comes from the non-differentiable sampling operation. Most of existing methods follow the research line of probability modeling, and process the non-differentiable sampling operation by two different methods: REINFORCE~\cite{DBLP:conf/emnlp/KeHHZ19, DBLP:conf/emnlp/LiMSJRJ17, DBLP:conf/nips/NorouziBCJSWS16} and continuous relaxations~\cite{DBLP:conf/icml/ZhangGFCHSC17, DBLP:conf/nips/ChenDTZGSZWZC18}. REINFORCE makes use of rewards from discriminators to update generators. This method is an unbiased but high variance estimator~\cite{DBLP:journals/corr/abs-2011-13527}. A number of methods are proposed to give more reasonable and stable rewards~\cite{DBLP:conf/aaai/GuoLCZYW18}, such as: obtaining rewards with Monte Carlo Search~\cite{DBLP:conf/aaai/YuZWY17}, changing the discriminator into a ranker~\cite{DBLP:conf/nips/LinLHSZ17} and training the generator by filling blank~\cite{DBLP:conf/iclr/FedusGD18}. Continuous relaxations like Gumbel-Softmax trick~
\cite{DBLP:conf/iclr/MaddisonMT17, DBLP:conf/iclr/JangGP17} allow gradients to pass through directly to generators. It is a low-variance but bias estimator, so more complex model structures like Relation Networks~\cite{DBLP:conf/nips/SantoroRBMPBL17} are adopted to provide a more informative signal~\cite{DBLP:conf/iclr/NieNP19}.

The methods mentioned above all rely on MLE pre-training. Even so, their performance still remains gaps comparing with MLE~\cite{DBLP:conf/iclr/CacciaCFLPC20}. Recently. researchers try to build language GANs without MLE pre-training~\cite{DBLP:journals/corr/abs-2011-13527, DBLP:conf/nips/dAutumeMRR19}. These attempts are comparable with MLE with the help of pre-trained embeddings which are learned from additional large scale corpus, so strictly speaking they are still based on pre-training techniques.

The dependence on pre-training techniques reveal the inherent limitations in existing language GANs. In contrast, our focus is on a promising line of Representation Modeling Methods (RMMs). Despite that previous attempts on RMMs fail to show is as an effective approach~\cite{DBLP:conf/nips/dAutumeMRR19} or can only generate sentences that are poor in both quality and diversity~\cite{kumar2020endtoend}.

\section{Model}
In this section, we firstly introduce the structure of InitalGAN. Next, we present dropout sampling and fully normalized LSTM to tackle invalid sampling method and unhealthy gradients, respectively. After that, we introduce the training objectives of InitalGAN.

\begin{figure}[htbp]
  \centerline{\includegraphics[scale=0.26]{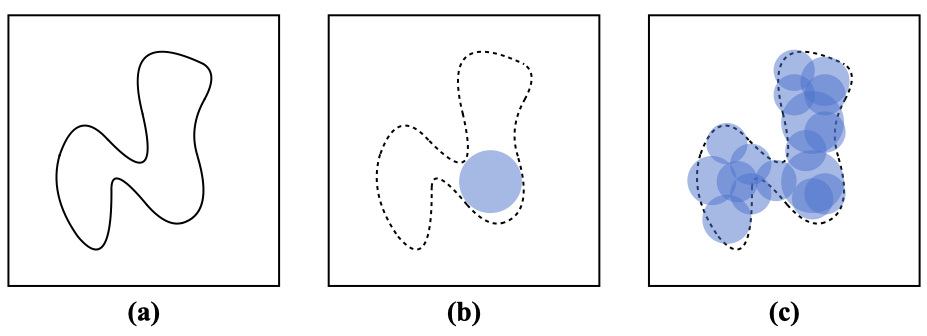}}
  \caption{Effects of Dropout Sampling. (a) The real distribution. (b) The distribution learned by a sub-model (blue area). (c) The distribution learned by the complete model.}
  \label{fig:dropout_sampling}
\end{figure}

\subsection{Model Structure}
The structure of InitialGAN is shown in Figure~\ref{fig:structure}. There are three models in InitialGAN: aligner, generator and discriminator. The aligner is based on the encoder in Transformer~\cite{DBLP:conf/nips/VaswaniSPUJGKP17}. It needs to transform words to representations. For the discriminator, we train it to identify whether a specific representation in the $t$-th timestep is from the aligner or the generator based on the previous $(t-1)$ representations:
\begin{equation*}
  \begin{aligned}
    \mathbf{c}_t = D_\psi(\mathbf{r}_t | \mathbf{r}_{t-1}, ..., r_1)
  \end{aligned}
  \label{eq-netD}
\end{equation*}
where $\mathbf{r}_t$ is the $t$-the representation from the aligner or the generator. For the generator, we use latent input $\mathbf{z}_t$ to generate representations and their corresponding words:
\begin{align}
    &\mathbf{r}_g^{(t)} = G_\varphi(\mathbf{z}_t | \mathbf{z}_{t-1}, ..., \mathbf{z}_0) \\
    &\hat{x}_t = F_{LT}(\mathbf{r}_g^{(t)})
    \label{eq-netG}
\end{align}
where $\mathbf{r}_g^{(t)}$ is the $t$-th representation and $F_{LT}(\cdot)$ is the transformation layer from the aligner. Its parameters are fixed during the training of the generator. In the following, we elaborate the details about the two important techniques in InitalGAN: Dropout Sampling and Fully Normalized LSTM.

\subsection{Dropout Sampling}

An effective sampling method plays a crucial role in training GANs. Once we choose to model representations, we can no longer sample words from word probabilities directly. Although we can sample results by feeding random noise into generators as a part of input, previous work~\cite{DBLP:conf/iclr/RazaviOPV19, DBLP:conf/acl/ZhuBLMLW20} shows that generators with autoregressive structures tend to ignore those additional input. As a result, these generators will suffer from the mode collapse problem~\cite{DBLP:conf/icml/ArjovskyCB17,DBLP:conf/nips/HeuselRUNH17} and give samples in high similarities. In this time, the discriminator can easily identify fake samples by remembering these highly similar data, and prompt generated samples to rotate between the different modes. Thus, the generator can only learn a small subset of the real distributions instead of the whole one.

Ever since Dropout~\cite{DBLP:journals/jmlr/SrivastavaHKSS14} was introduced in 2014, it has been widely used in training neural networks. Previous work~\cite{DBLP:conf/cvpr/IsolaZZE17} also adopts dropout as random noises in image GANs. Using dropout in both training and inference as a sampling method is extremely suitable in representation modeling methods because it provides a non-negligible random factor. During training, the hidden representations in the generator will be randomly masked. Under the guidance of the discriminator, the generator will learn to use different combinations of the values in hidden representations to obtain results following real distributions. During inference, the generator has consistent masking strategy to ensure the quality of generated results.

However, the distribution provided by dropout sampling is decided by the input which is always in the form of trainable embeddings. It leads the distribution to keep changing during training and may cause the training process to be unstable. Thus, we concatenate the input with random noise to increase the robustness of the model. The complete method is:
\begin{equation*}
  \mathbf{z}_t = Dropout(E(\hat{x}_{t-1}) \oplus \mathbf{\epsilon}, \rho)
\end{equation*}
where $\mathbf{z}_t$ is the $t$-th latent variable, $\hat{x}_{t-1}$ is the word generated in the last timestep, ${E}(\cdot)$ is a function to transform words into embeddings, $\epsilon$ is random noise sampled from a pre-defined distribution and $\rho$ is the dropout rate. Dropout can be viewed as the selection of sub-models. Given a real distribution in Figure~\ref{fig:dropout_sampling} (a), even though each sub-model may still suffer from mode collapse as shown in Figure~\ref{fig:dropout_sampling} (b), different sub-models can cover different modes. Hence the distribution given by the complete model is closer to the real data distribution (as shown in Figure~\ref{fig:dropout_sampling} (c)).

Dropout sampling will slow down the convergence of generators, since the parameters are updated less frequently. To speed up the training process, we propose to use imbalanced batch size. Suppose $bs_d$ is the batch size of discriminator's training, setting the batch size of the generator's training as $bs_g = bs_d / (1-\rho)$ can bridge the gap in update frequency.

\subsection{Fully Normalized LSTM} 

When building representation modeling based language GANs, generators update their parameters based on the gradients from discriminators. Compared with other sequence models~\cite{DBLP:conf/naacl/PetersNIGCLZ18}, representation modeling methods place higher demands on healthy gradients, since they need to stack more layers to build discriminators and generators.

In a language GAN, a generator makes predictions based on previous output in both training and inference. It limits the use of Transformer~\cite{DBLP:conf/nips/VaswaniSPUJGKP17} whose computational speed is extremely slow without parallel computation. When generating the $t$-th word, Transformer needs to calculate attention weights for the previous $t-1$ words, and it is quadratic complexity over the input length. Consequently, LSTM~\cite{DBLP:journals/neco/HochreiterS97} is more popular in language GANs~\cite{DBLP:conf/nips/dAutumeMRR19}. To obtain the output, it only needs to consider the hidden state from the last timestep and the current input, so it has a constant fast computational speed. 

When using LSTM to build a representation modeling method, however, we need to care about possible gradient vanishment among both different timesteps and different layers. This time, the gated mechanism in LSTM cuts both ways. Although the grated mechanism can relieve gradient vanishment among different timesteps, the hidden states of LSTM needs to multiply with the results from the output gate, whose values are between $(0, 1)$. Thus, the gradients from the previous layers will be inevitably narrowed.

The problem gets worse when we stack several LSTM layers to build the generator and the discriminator. Unhealthy gradients will exist throughout the whole training process and affect model performance directly. Thus, we need to find a method to relieve the gradient vanishment problem.

Layer normalization~\cite{DBLP:journals/corr/BaKH16} is a widely used technique in neural networks~\cite{DBLP:conf/nips/VaswaniSPUJGKP17}. Previous work~\cite{DBLP:conf/nips/Xu0ZZL19} shows that layer normalization helps stabilize training by reducing the variance of gradients. We further find that layer normalization has potential to relieve gradient vanishment. A common understanding is that it can shift and scale input into a more reasonable interval to avoid the interval whose gradients are small. However, layer normalization does more than that. According to our analyses, it provides an addition term to augment gradients when the deviation of the normalized term is smaller than 1, as stated in Theorem 1 below.

\paragraph*{Theorem 1} \textit{Suppose $\mathbf{y}_{l+1}=F_l(\mathbf{y}_l)$ is the $l$-th layer in a model, $\mathbf{y}_l$ is the input of the $l$-th layer and also the output from the $(l-1)$-th layer. Adopting layer normalization in the input (i.e., $\mathbf{y}_{l+1}=F_l(LN(\mathbf{y}_l))$) provides an addition term when calculating the partial derivative of $\mathbf{y}_{l+1}$ with respect to $\mathbf{y}_l$. This term can augment the gradients when the deviation of $\mathbf{y}_l$ is smaller than 1.}

In LSTM, the output of LSTM is the element-wise product of the output gate (which is in $(0, 1)$) and cell state (which is in $(-1, 1)$). Its deviation must be smaller than 1, which can meet the conditions of Theorem 1. Thus, we have the following:

\paragraph*{Corollary 1} \textit{Adopting layer normalization when calculating the hidden state in LSTM provides a scaler factor to augment the gradients, thereby mitigating the gradient vanishment between different layers in LSTM.}

The proofs of Theorem 1 and Corollary 1 can be found in Appendix~\ref{appendix_theorem_1} and~\ref{appendix_corollary_1}, respectively. Based on our analyses, we propose a fully normalized LSTM as follows:
\begin{equation}
  \begin{aligned}
    \begin{pmatrix}
      \mathbf{f}_t \\
      \mathbf{i}_t \\
      \mathbf{o}_t \\
      \mathbf{\hat{c}}_t
    \end{pmatrix} &=LN(\mathbf{W}_h \mathbf{h}_{t-1}) + LN(\mathbf{W}_x \mathtt{x}_t) + \mathbf{b} \\
    \mathbf{c}_t &= LN(F_s(\mathbf{f}_t) \circ \mathbf{c}_{t-1} + F_s(\mathbf{i}_t) \circ F_h(\mathbf{\hat{c}}_t)) \\
    \mathbf{h}_t &= LN(F_s(\mathbf{o}_t) \circ F_h(\mathbf{c}_t)) \label{eq-fullylnlstm}
  \end{aligned}
\end{equation}
where $\circ$ is the element-wise product, $F_s(\cdot)$ is sigmoid function and $F_h(\cdot)$ is tanh. The fully normalized LSTM is based on LayerNorm LSTM~\cite{DBLP:journals/corr/BaKH16} whose effectiveness has been shown by various tasks~\cite{DBLP:conf/nips/dAutumeMRR19, DBLP:conf/nips/HouZKGQL19}. The main difference is that the fully normalized LSTM adopts additional layer normalization when calculating the hidden state, so as to offset the influence from the output gate. Thus, it can obtain both strong sequence modeling capabilities and healthier gradients. In practice, we adopt fully normalized LSTM to construct both the discriminator and the generator to obtain healthier gradients

\subsection{Training Objective}

The training objective of the aligner is based on the loss function of Variational Autoencoder (VAE)~\cite{DBLP:journals/corr/KingmaW13}, so the aligner can map word $x_i$ into a distribution $\mathcal{N}(\mu_{x_i}, \sigma_{x_i}^2)$. The vector sampled from $\mathcal{N}(\mu_{x_i}, \sigma_{x_i}^2)$ is transformed back into words with a linear transformation $F_{LT}$. The distribution $\mathcal{N}(\mu_{x_i}, \sigma_{x_i}^2)$ describes a region in the space as representation rather than a point. It can increase the robustness during generation, since $F_{LT}$ can map representations with minor errors into correct words.

\begin{algorithm}
  \caption{Training of InitialGAN}\label{alg-training}
  \begin{algorithmic}[1]
  \REQUIRE Initial parameters of Variational Aligner $\phi_0$, Discriminator $\psi_0$, and Generator $\varphi_0$. Batch size $m$ and imbalanced batch size $m^\prime$. \\
  \STATE  // Training the aligner \\
  \WHILE{$\phi$ has not converged}
  \FOR{$i = 1\rightarrow m$}
  \STATE Sample real data $x \sim \mathbb{P}_x $\\
  \STATE $\mu_{x}, \sigma^2_{x} \leftarrow A_\phi({x})$\\
  \STATE $\hat{\epsilon} \sim \mathcal{N}(0,1)$ \\
  \STATE $\mathbf{\hat{z}} \leftarrow \hat{\epsilon} \circ \sigma_{{x}} + \mu_{{x}}$ \\
  \STATE $L^i_A \leftarrow -\mathbb{E}_{\mathbf{\hat{z}} \sim q(\mathbf{\hat{z}} |{x})}(logp({x} | \mathbf{\hat{z}})) + KL(q(\mathbf{\hat{z}} |{x})||p(\mathbf{\hat{z}})) + \lambda_a log(\sigma^2_{{x}})$
  \STATE  $\phi \leftarrow AdamW(\nabla_\phi \frac{1}{m}\sum_{i=1}^m L^i_A, \phi)$
  \ENDFOR
  \ENDWHILE
  \STATE // Training the discriminator and the generator \\
  \WHILE{$\varphi$ has not converged}
  \FOR{$i = 1 \rightarrow m$}
  \STATE Sample real data $x \sim \mathbb{P}_x $, latent variable $\mathbf{z} \sim \mathbb{P}_z$, random number $\epsilon^\prime \sim Uniform[0, 1]$. \\
  \STATE  $\mu_{{x}}, \sigma^2_{{x}} \leftarrow A_\phi({x})$\\
  \STATE  $\mathbf{r}_d \leftarrow \mu_{{x}}$ \\
  \STATE  $\mathbf{r}_g \leftarrow G_\varphi(\mathbf{z})$ \\
  \STATE  $\mathbf{r}_m \leftarrow \epsilon^\prime \cdot \mathbf{r}_d + (1 - \epsilon^\prime) \cdot \mathbf{r}_g$ \\
  \STATE  $L_D^i \leftarrow D_\psi (\mathbf{r}_g) -  D_\psi (\mathbf{r}_d) + \lambda_d\cdot R$
  \ENDFOR
  \STATE $\psi \leftarrow AdamW(\nabla_\psi \frac{1}{m}\sum_{i=1}^m L^i_D, \psi) $
  \STATE Sample a batch of latent variables $\{\mathbf{z}^{(i)}\}^{m^\prime}_{i=0} \sim \mathbb{P}_z$
  \STATE $\varphi \leftarrow Adam(\nabla_\varphi - \frac{1}{m^\prime}\sum_{i=1}^{m^\prime} D_\psi(G_\varphi(\mathbf{z}^{(i)})), \varphi)$
  \ENDWHILE
  \end{algorithmic}
\end{algorithm}

However, the aligner trained via the VAE objective tends to assign large regions to high frequency words, while the regions of low frequency words are extremely small. It brings difficulties to generate low frequency words, since small errors may lead the representations to lie in the region of other words. Thus, we propose a new training objective to tackle this problem:
\begin{equation}
  \begin{aligned}
  L_{A} &=-\mathbb{E}_{\mathbf{\hat{z}}\sim q(\mathbf{\hat{z}}|x_i)}(logp(x_i|\mathbf{\hat{z}})) + KL(q(\mathbf{\hat{z}}|x_i)||p(\mathbf{\hat{z}}))  \\
  & \qquad + \lambda_a log(\sigma^2_{x_i}) \label{eq-loss_vp}
  \end{aligned}
\end{equation}
where $\sigma^2_{x_i}$ is the variance of word $x_i$ and $\lambda_a$ is a hyperparameter. The first two terms consist of the original objective in VAE. The last term provides a penalization to the variance, which controls the size of the representation region directly. This objective can help keep the regions of different words in similar size by giving more penalization to the variance of high frequency words. 

We train the aligner based on the strategy of BERT~\cite{DBLP:conf/naacl/DevlinCLT19}. More specifically, 15\% of words in training data will be randomly selected as the ones to be predicted. Among the replacing words, 80\% of them are replaced with [MASK] token, 10\% of them are replaced with random tokens, and 10\% of them are unchanged. In the training objective of the aligner, we propose to add an additional term to penalize $\sigma^2_{x_i}$. Vanilla VAE also needs to calculate $\sigma^2_{x_i}$ to update the KL divergence term in the objective, so there is nearly no additional computational cost in the new training objective. After the training of the aligner is finished, all its parameters are fixed and the transformation layer which transforms representations back into words will be shared with the generator.

Both the discriminator and the generator are constructed based on the fully normalized LSTM. Dropout sampling is adopted in both training and inference stage of the generator. The generator uses the same linear transformation $F_{LT}$ in the aligner to transform representations back into words. We adopt Wasserstein distance~\cite{DBLP:conf/icml/ArjovskyCB17} as the training objective, and use Lipschitz penalty~\cite{DBLP:conf/iclr/PetzkaFL18} to regularize the discriminator. The loss functions of the discriminator $L_D$ and the generator $L_G$ are:
\begin{align}
    &L_D = - \mathbb{E}_{\mathbf{r}_d\sim \mathbb{P}_d}[D(\mathbf{r}_d)] + \mathbb{E}_{z \sim \mathbb{P}_{z}}[D(G(\mathbf{z}))] + \lambda_d R \label{eq-wgan-1} \\
    &L_G = - \mathbb{E}_{z \sim \mathbb{P}_{z}}[D(G(\mathbf{z}))] \label{eq-wgan-2}
\end{align}
where $\mathbf{z}$ is from dropout sampling, $\mathbf{r}_m$ is sampled uniformly along straight lines between pairs of representations from the aligner and the generator, $R=\mathbb{E}_{\mathbf{r}_m \sim \mathbb{P}_m}[(max\{0, ||\nabla D(\mathbf{r}_m)||_2-1\})^2]$, and $\mathbf{r}_d$ is the word representation from the aligner, which is $\mu_{x_i}$ in this work. The full training process of InitialGAN is described in Algorithm~\ref{alg-training}.

\begin{figure*}[htbp]
  \centerline{\includegraphics[scale=0.29]{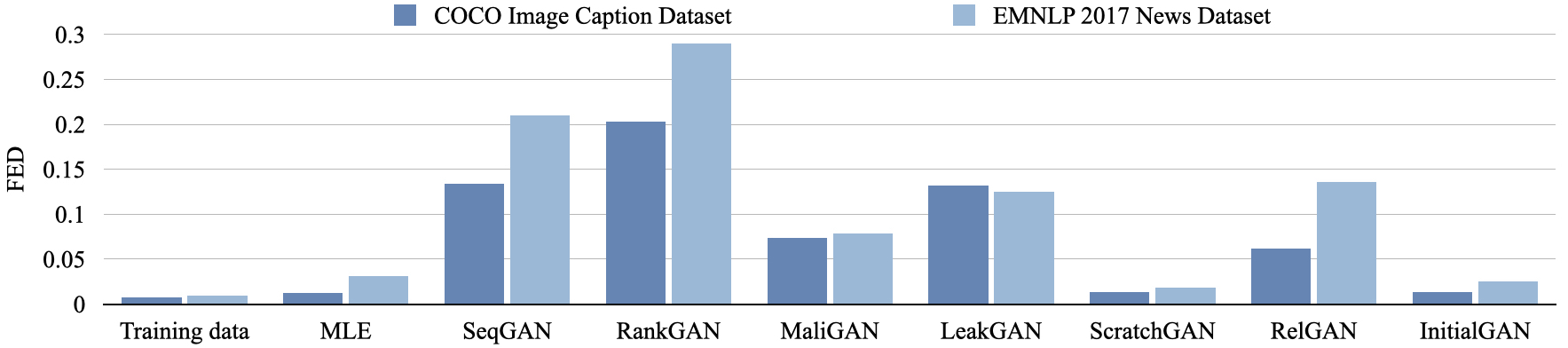}}
  \caption{Evaluation Results in Fr{\'{e}}chet Embedding Distance (FED) on COCO Caption Dataset and EMNLP 2017 News Dataset. Lower is Better.}
  \label{fig:fed-coco}
\end{figure*}

\section{Experiment}
\label{sec-exp}

In this section, we introduce the evaluation metrics, datasets and compared models. After that, we provide the experimental results with our analyses. In the experiment, we firstly demonstrate the effectiveness of InitialGAN in both short sentence generation and long sentence generation. Then, we show the reliance of existing models on pre-trained techniques. After that, we conduct ablation study to show the effectiveness of dropout sampling and fully normalized LSTM. Last but not least, models are further compared from the training stability and sentence length distribution. 

\subsection{Evaluation Metrics}
\label{sec-eva_metrics}

For the token level metrics, we use \textbf{BLEU}~\cite{DBLP:conf/acl/PapineniRWZ02} to evaluate fluency and \textbf{Self-BLEU}~\cite{DBLP:conf/sigir/ZhuLZGZWY18} to evaluate diversity. Caccia et al.~\cite{DBLP:conf/iclr/CacciaCFLPC20} propose to tune the temperature in the softmax to draw a curve so as to evaluate fluency and diversity together. However, InitialGAN models word representations, and does not use softmax to obtain specific word probabilities. Thus, we can not use this method to evaluate the performance. Instead, \textbf{Inverse BLEU} is a good choice for evaluating the overall performance in both fluency and diversity. Inverse BLEU uses sentences in test sets as inference and generated sentences as references. Sentences in test sets are fluency and diverse, so the generated sentences can get high Inverse BLEU only when they have good performance in terms of both aspects. When calculating token level metrics, all of them are calculated up to 5 grams and the size of each set is set to be 5,000.

For the embedding level metrics, we use \textbf{Fr{\'{e}}chet Embedding Distance (FED)}~\cite{DBLP:conf/nips/dAutumeMRR19} which is identical with Fr{\'{e}}chet inception distance (FID)~\cite{DBLP:conf/nips/HeuselRUNH17} except for the encoding model. Although it can evaluate the global similarity of two distributions, previous work~\cite{DBLP:journals/corr/abs-2011-13527, DBLP:conf/nips/dAutumeMRR19} shows that its values are extremely small, and it is not sensitive to the change of sample quality. When models get similar FED, it does not mean that they get close performance. To further identify the differences of compared models, we propose a new metric, \textbf{Least Coverage Rate (LCR)}, which is calculated as follows:
\begin{equation}
  \begin{aligned}
    &S_{ij} = Sim(\mathbf{E}(\mathbf{x}^a_i), \mathbf{E}(\mathbf{x}^b_j)) \\
    &\mathbf{R}_a = \frac{1}{n} \sum_{i=1}^n \delta(\sum_{j=1}^m S_{ij} \ge \tau) ,\ \mathbf{R}_b = \frac{1}{m} \sum_{j=1}^m \delta(\sum_{i=1}^n S_{ij} \ge \tau)  \\
    &LCR(\mathbf{X}_a, \mathbf{X}_b) = min(\mathbf{R}_a, \mathbf{R}_b) \label{eq-lcr}
  \end{aligned}
\end{equation}
where $\mathbf{x}^a_i$ and $\mathbf{x}^b_i$ are the $i$-th and $j$-th sentences from sentence sets $\mathbf{X}_a$ and $\mathbf{X}_b$, respectively. $\mathbf{E}(\cdot)$ is the model to transform sentences into embeddings, $\tau$ is a hyperparameter, $Sim(\cdot)$ is a similarity function and $\delta(\cdot)$ is a function which returns 1 if input is higher than 0, and 0 for others. 

$\mathbf{R}_a$ and $\mathbf{R}_b$ are the coverage rates of $\mathbf{X}_a$ and $\mathbf{X}_b$, respectively. LCR uses minimum value between $\mathbf{R}_a$ and $\mathbf{R}_b$ as the final result, so it is aware of two common cases: 1) a majority of generated samples are out of real distribution; 2) generated samples are real-like but in high similarities. LCR compares the similarity of two distributions in a fine-grained level, hence it is more sensitive to the changes in samples. More explanations and comparisons between FED and LCR can be found in Appendix~\ref{appendix_lcr}.

When adopting these two embedding level metrics, we select 10,000 sentences in each set and use Universal Sentence Encoder~\cite{DBLP:conf/emnlp/CerYKHLJCGYTSK18}\footnote{\url{https://tfhub.dev/google/universal-sentence-encoder/4}} to encode them into embeddings. For the similarity function in LCR, we use cosine similarity suggested by the Universal Sentence Encoder~\cite{DBLP:conf/emnlp/CerYKHLJCGYTSK18}.

\begin{figure}[htbp]
  \centerline{\includegraphics[scale=0.24]{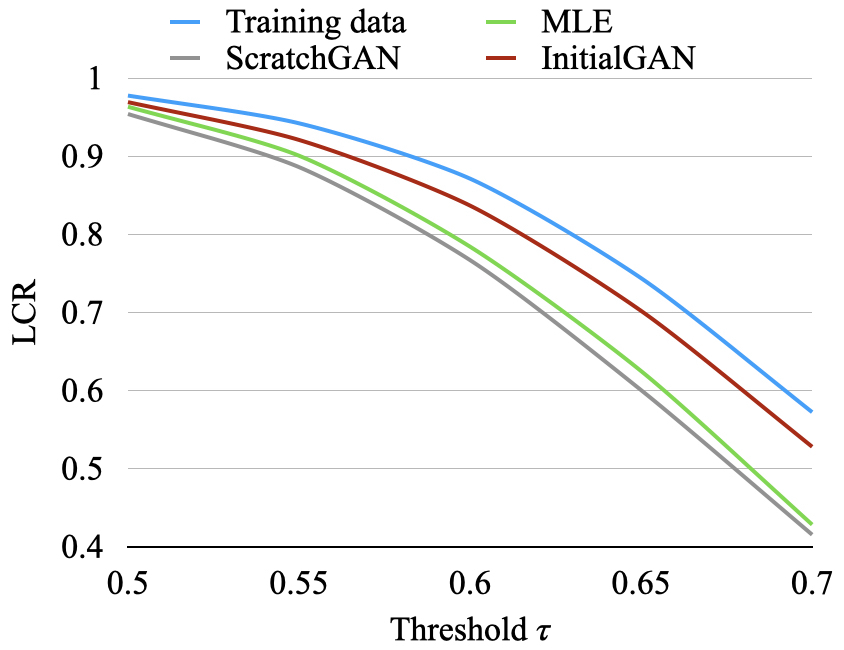}}
  \caption{LCR with different $\tau$ on the COCO Dataset. Higher is Better.}
  \label{fig:lcr-coco}
\end{figure}

\begin{figure}[htbp]
  \centerline{\includegraphics[scale=0.24]{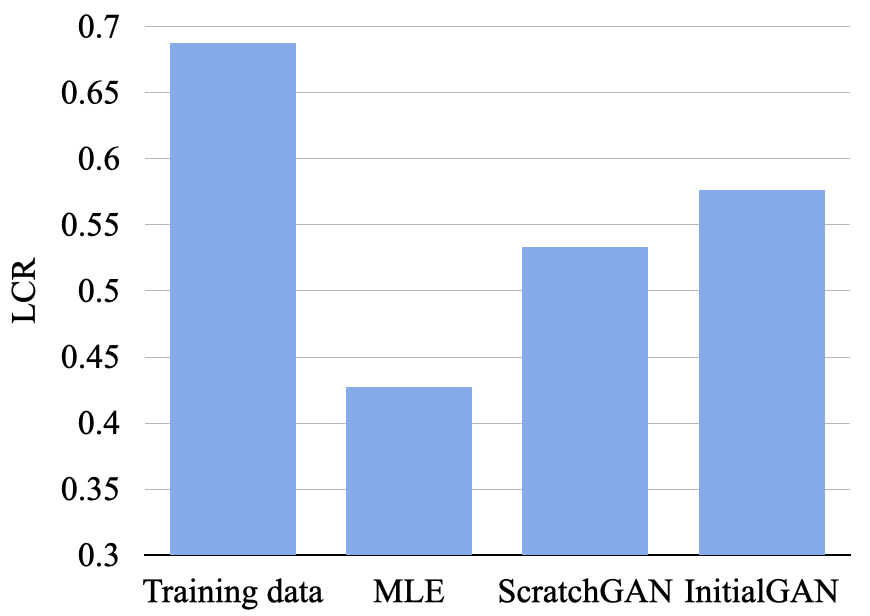}}
  \caption{LCR on EMNLP 2017 News Dataset ($\tau=0.45$). Higher is Better.}
  \label{fig:lcr-emnlp}
\end{figure}

\subsection{Experiment Setup}
\label{sec-exp_setup}

We use two datasets in our experiment: COCO Image Caption Dataset~\cite{DBLP:conf/eccv/LinMBHPRDZ14}~\footnote{\url{https://cocodataset.org}} and EMNLP 2017 News Dataset~\footnote{\url{http://www.statmt.org/wmt17/}}. For the COCO Image Caption Dataset, we choose 50,000 sentences as training set. For the EMNLP 2017 News Dataset, we choose 200,000 sentences as training set.

We compare the performance of InitialGAN with MLE and other language GANs. For REINFORCE methods, we choose SeqGAN~\cite{DBLP:conf/aaai/YuZWY17}, RankGAN~\cite{DBLP:conf/nips/LinLHSZ17}, MaliGAN~\cite{DBLP:journals/corr/CheLZHLSB17}, LeakGAN~\cite{DBLP:conf/aaai/GuoLCZYW18} and ScratchGAN~\cite{DBLP:conf/nips/dAutumeMRR19}. For continuous relaxation methods, we choose RelGAN~\cite{DBLP:conf/iclr/NieNP19}. All these methods except for ScratchGAN rely on MLE pre-training, while ScratchGAN is based on pre-trained embeddings. InitialGAN is the only language GAN whose parameters are initialized completely randomly. More experimental details can be found in Appendix~\ref{appendix-exp_details}.

\subsection{Experimental Results}
\label{sec-exp_result}

Figure~\ref{fig:fed-coco} reports FED of different models. On the COCO Image Caption Dataset, MLE, ScratchGAN and InitialGAN can significantly outperform other compared models; the differences among these three models are not clear. On the EMNLP 2017 News Dataset, ScratchGAN and InitialGAN slightly outperform MLE, though the gaps between these three models are still limited. FED can not effectively capture the change of data qualities.

To better evaluate the performance among these three models, we further compare their performance in LCR. The results are shown in Figures~\ref{fig:lcr-coco} and~\ref{fig:lcr-emnlp}. We explore the effectiveness of the threshold $\tau$ in LCR on Image COCO Caption Dataset. The results are demonstrated in Figure~\ref{fig:lcr-coco}. Although the values change significantly with different $\tau$, the rankings of different models are kept when $\tau$ is in a reasonable interval. According to Figure~\ref{fig:lcr-coco}, ScratchGAN is slightly inferior to MLE while InitialGAN can outperform both models. Figure~\ref{fig:lcr-emnlp} shows LCR on EMNLP 2017 News Dataset. InitialGAN gets the highest LCR among all three models. Unlike on Image COCO Caption Dataset, ScratchGAN outperforms MLE on this dataset. EMNLP 2017 News Dataset consists of long sentences and its distribution is more complicated. The exposure bias problem in MLE is more likely to happen, which leads to the poor performance on this dataset.

\begin{table}[]
  \centering
  \caption{Evaluation Results of Token Level Metrics on Image COCO Caption Dataset (S-BLEU: Self-BLEU, I. BLEU: Inverse BLEU)}
  \begin{tabular}{|c|c|c|c|}
  \hline
  \textbf{Model}& \textbf{BLEU} & \textbf{S-BLEU} & \textbf{I. BLEU}   \\ \hline \hline
  Training Data & 34.99 &  34.80  & 35.36   \\ \hline \hline
  MLE           & 32.59 &  37.15  & 32.03   \\ \hline
  SeqGAN        & 34.68 &  69.85  & 22.34   \\ \hline
  RankGAN       & 37.32 &  73.30  & 22.10   \\ \hline
  MaliGAN       & 26.49 &  53.47  & 25.95   \\ \hline
  LeakGAN       & 33.14 &  56.88  & 29.43   \\ \hline
  ScratchGAN    & 30.98 &  35.72  & 30.76   \\ \hline
  RelGAN        & 54.04 &  73.70  & 29.53   \\ \hline \hline
  InitialGAN    & 34.87 &  39.06  & \textbf{33.06}  \\ \hline
  \end{tabular}
  \label{table:token-metric-coco}
\end{table}

\begin{table}[]
  \centering
  \caption{Evaluation Results of Token Level Metrics on EMNLP 2017 News Dataset (S-BLEU: Self-BLEU, I. BLEU: Inverse BLEU)}
  \begin{tabular}{|c|c|c|c|}
  \hline
  \textbf{Model}& \textbf{BLEU} & \textbf{S-BLEU} & \textbf{I. BLEU}   \\ \hline \hline
  Training Data & 20.50 &  20.47  & 20.62   \\ \hline \hline
  MLE           & 16.66 &  17.21  & 16.97   \\ \hline
  SeqGAN        & 9.01  &  27.89  & 9.90   \\ \hline
  RankGAN       & 10.35 &  56.77  & 10.37   \\ \hline
  MaliGAN       & 12.23 &  21.34  & 13.11   \\ \hline
  LeakGAN       & 27.61 &  50.55  & 11.59   \\ \hline
  ScratchGAN    & 17.54 &  19.04  & 17.19   \\ \hline
  RelGAN        & 30.95 &  57.48  & 14.74   \\ \hline  \hline
  InitialGAN    & 19.40 &  23.74  & \textbf{17.74}  \\ \hline
  \end{tabular}
  \label{table:token-metric-emnlp}
\end{table}

The evaluation results in token level evaluation metrics are shown in Tables~\ref{table:token-metric-coco} and~\ref{table:token-metric-emnlp}. We firstly analyze the results on COCO Image Caption Dataset. Except for ScratchGAN and InitialGAN, most of language GANs tend to get very high Self-BLEU. Sentences generated by these models have high similarities, which indicates the mode collapse problem in these models. ScratchGAN can tackle this problem and get better result in Inverse BLEU. However, compared with MLE, it still has a gap even with the help of pre-trained embeddings. InitialGAN is the only language GAN which can outperform MLE when considering fluency and diversity together.

\begin{figure}[htbp]
  \centerline{\includegraphics[scale=0.23]{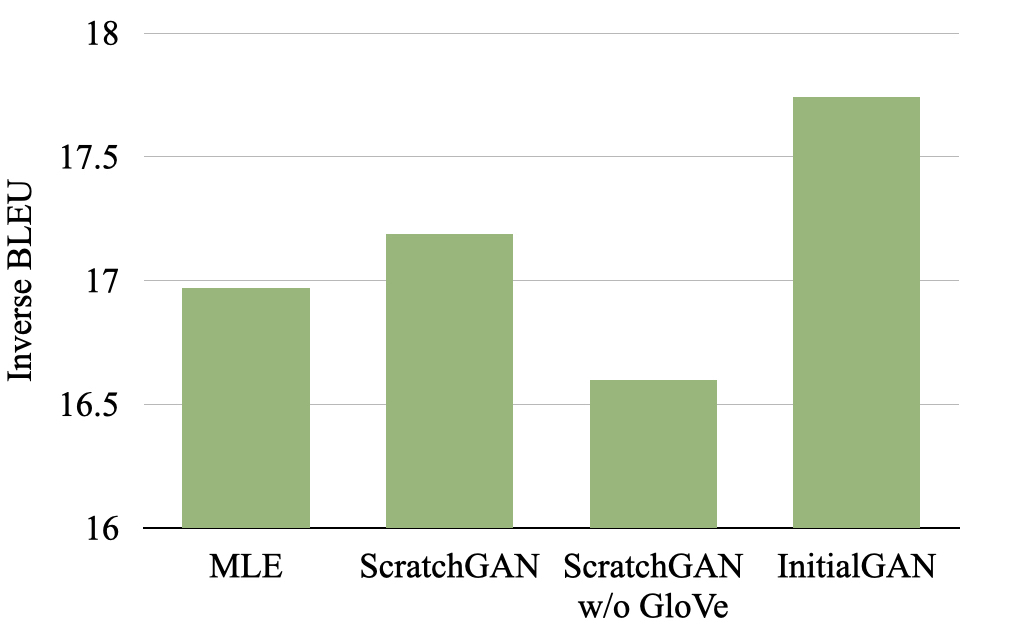}}
  \caption{Ablation Study of ScartchGAN on EMNLP 2017 News Dataset.}
  \label{fig:ablation-1}
\end{figure}

\begin{figure}[htbp]
  \centerline{\includegraphics[scale=0.24]{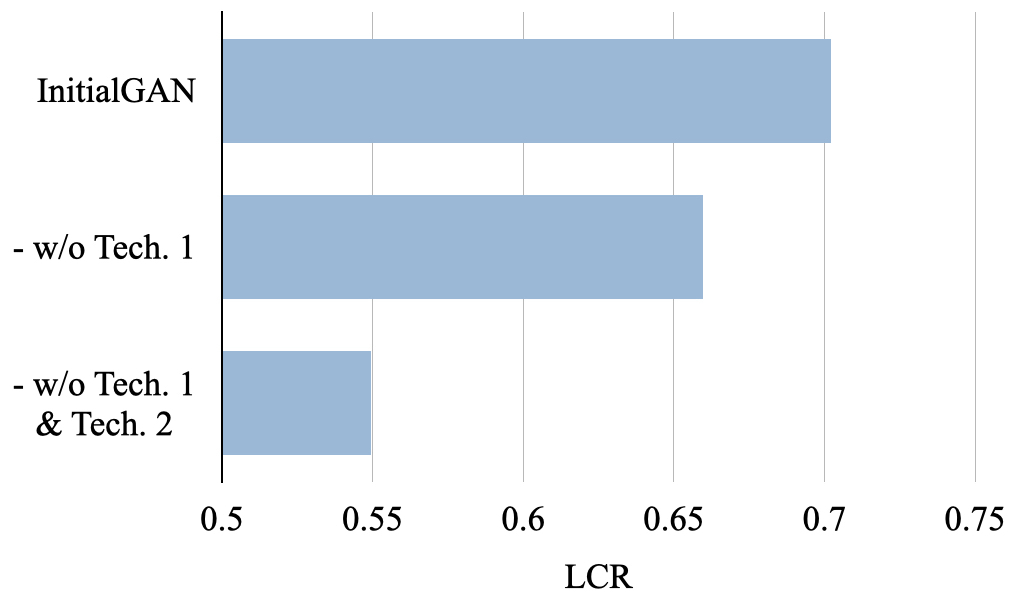}}
  \caption{Ablation Study of InitialGAN on COCO Dataset. Tech. 1: Dropout Sampling. Tech. 2: Fully Normalized LSTM.}
  \label{fig:ablation-2}
\end{figure}

Similar results can be found on EMNLP 2017 News Dataset. The difference is that ScratchGAN can slightly outperform MLE in terms of Inverse BLEU. It is consistent with the results in LCR. When generating long sentences, MLE is more likely to meet exposure bias, so MLE gets lower BLEU, which means these sentences are lack of local consistency. Besides, the Self-BLEU shows that sentences generated by MLE are more diverse than training data. A number of generated sentences are out of the real distribution. It explains why MLE only get unsatisfactory performance on this dataset.

We further explore the performance of ScartchGAN without pre-trained embeddings, and show the results in Figure~\ref{fig:ablation-1}. Once we remove the pre-trained embeddings, ScartchGAN can no longer outperform MLE, and the gap between ScratchGAN and InitialGAN becomes larger. It reflects the dependence of ScartchGAN on pre-trained embeddings.

Existing language GANs highly rely on pre-training techniques to be comparable to MLE, while InitialGAN is the only language GAN, which can get better performance without using any pre-training techniques. It demonstrates the effectiveness of RMMs against high variance REINFORCE or biased continuous relaxation methods.

\begin{figure}[htbp]
  \centerline{\includegraphics[scale=0.25]{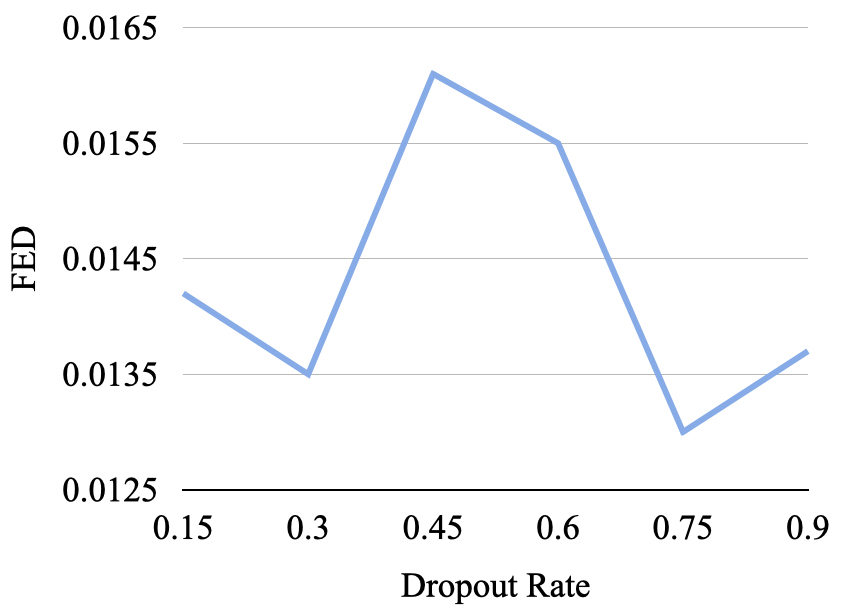}}
  \caption{Dropout Sampling with different dropout rates.}
  \label{fig:ablation-3}
\end{figure}

\begin{figure}[htbp]
  \centerline{\includegraphics[scale=0.25]{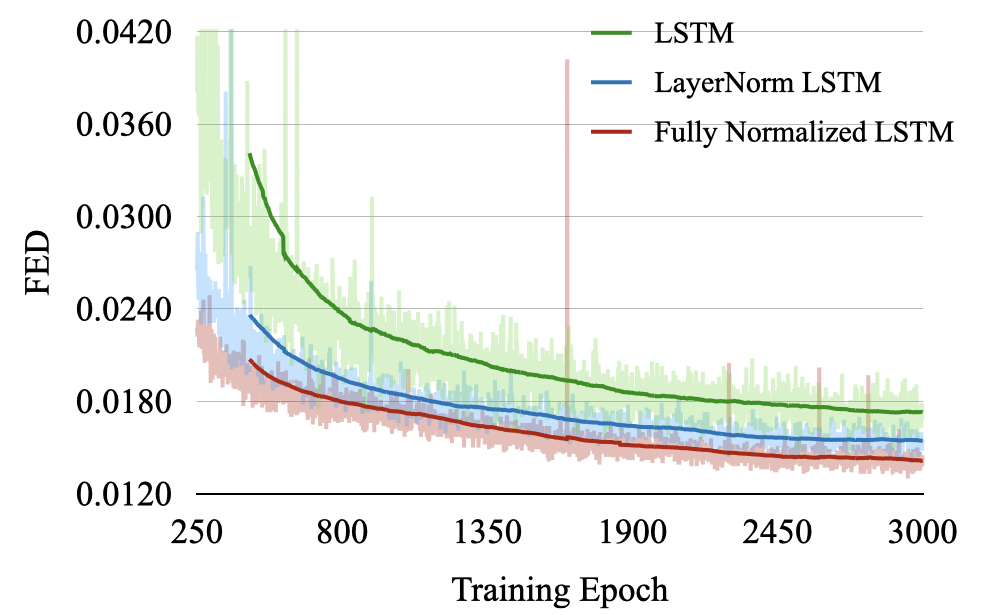}}
  \caption{ Comparisons of different LSTM.}
  \label{fig:ablation-4}
\end{figure}

Figure~\ref{fig:ablation-2} shows the experimental results of the ablation study about dropout sampling and fully normalized LSTM. Without dropout sampling, the LCR decreases a lot because of the mode collapse problem. In this time, the generated samples have extremely high similarities and its Self-BLEU is 71.88, which is much higher than the value of real data. The situation gets worse if we further remove the fully normalized LSTM. These results demonstrate the importance of dropout sampling and fully normalized LSTM to the performance of InitialGAN. Additional experiments on COCO Image Caption Dataset are conducted to further explore these two proposed techniques. The results are shown in Figure~\ref{fig:ablation-3} and Figure~\ref{fig:ablation-4}.

Figure~\ref{fig:ablation-3} shows the influence of dropout rate to FED. The curve shows a rough symmetry. We suppose it comes from the symmetry of combinations. Given a $d$-dimension vector, the number of possible combinations of masking $\rho \cdot d$ dimensions is the same as the number of masking $(1 - \rho) \cdot d$ dimensions ($\rho$ is the dropout rate). Figure~\ref{fig:ablation-4} shows the change of FED on the validation set during the training process. LayerNorm LSTM, the original combination of LSTM and layer normalization~\cite{DBLP:journals/corr/BaKH16}, can outperform LSTM, while our fully normalized LSTM can further speed up convergence and get better FED.

\begin{figure}[htbp]
  \centerline{\includegraphics[scale=0.25]{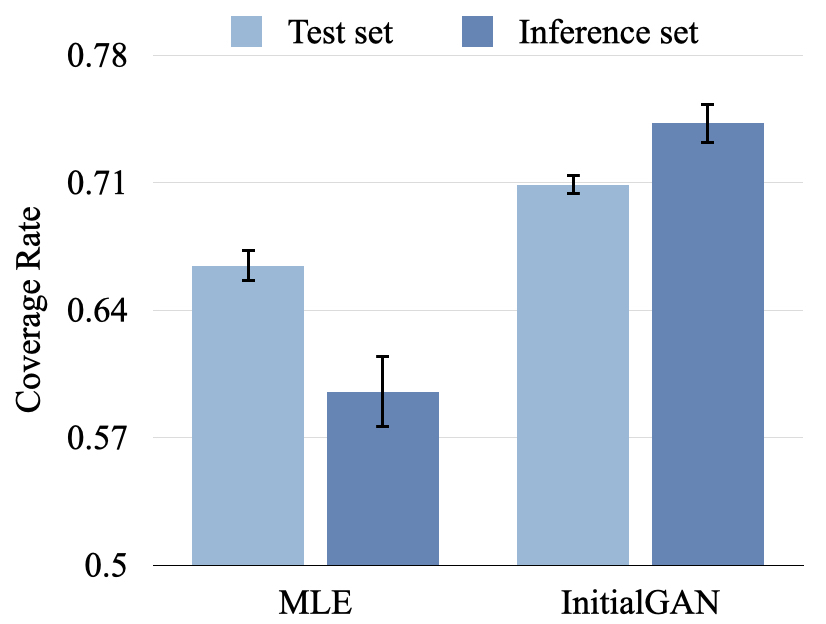}}
  \caption{Coverage Rate of MLE and InitialGAN ($\tau=0.65$).}
  \label{fig:additional-result-1-1}
\end{figure}

\begin{figure}[htbp]
  \centerline{\includegraphics[scale=0.25]{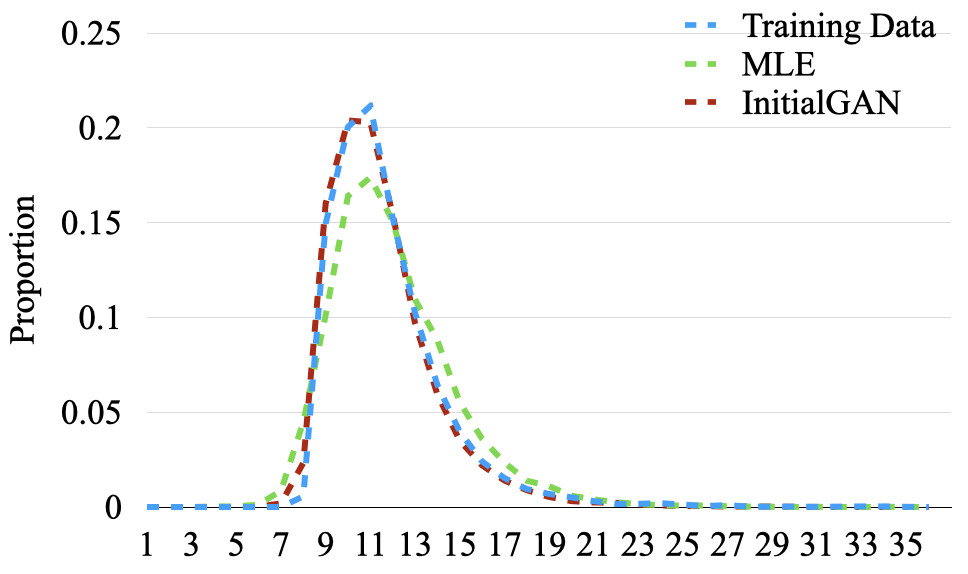}}
  \caption{Sentence length distribution.}
  \label{fig:additional-result-1-2}
\end{figure}

LCR uses the minimum values among two coverage rates as the final results, and analyzing these two coverage rates can also help us better understand the models (more explanations about it can be found in Appendix~\ref{appendix_lcr}). We train MLE and InitialGAN with different random seeds on COCO Image Caption Dataset, and show their average coverage rates with the standard deviations in Figure~\ref{fig:additional-result-1-1}. InitialGAN gets higher coverage rates and its standard deviation is slightly smaller than that of MLE. It shows InitialGAN can get consistently better result with different random seeds. Besides, MLE gets a lower coverage rate in the inference set. We regard the exposure bias as the cause leading MLE to generate sentences out of the real distribution, so its coverage rate of inference set is lower. InitialGAN gets higher coverage rate on the inference set, so further work is needed to relieve the mode collapse problem. We also show the sentence length distributions in Figure~\ref{fig:additional-result-1-2}. Compared with MLE, sentences generated by InitialGAN have a closer distribution to the training data. More experimental results and generated samples can be found in Appendix~\ref{appendix-exp_results} and~\ref{appendix-samples}, respectively.

\section{Conclusion}
\label{sec-conclusion}

In this work, we conduct an in-depth study about constructing language GANs based on representation modeling. We analyze two main problems which limit the performance of representation modeling methods: invalid sampling and unhealthy gradients. To tackle the invalid sampling, we introduce dropout sampling, a simple but effective method. For the unhealthy gradients, we conduct thorough analyses of layer normalization and present the fully normalized LSTM. Armed with these two techniques, we propose InitialGAN which is composed of three models, i.e., aligner, generator and discriminator. Different from existing language GANs which are based on pre-training techniques, all the parameters in InitalGAN are initialized randomly. Besides, we find that FED is not sensitive to the change of sample quality, so we propose Least Coverage Rate (LCR) to better identify the differences among different models. We conduct experiments on two widely used datasets, and the experimental results show that InitialGAN can outperform both MLE and other compared models. To the best of our knowledge, it is the first time that a language GAN can outperform MLE without using any pre-training techniques. This work also demonstrates that RMMs denote a promising research line for language GANs.

Although language GANs can tackle the exposure bias problem, their training speed is one of the most notable limitations which must be tackled. Language GANs need to use previously generated tokens as input in both training and inference, so its training speed can not be improved by parallel computation structures like Transformer. It limits the applications of current language GANs on large datasets. How to improve the training speed is a key problem that needs to be solved in future research.

\section*{Acknowledgments}
We are grateful to Hongzhi Zhao for his help in mathematical proofs. The research of this paper has been supported by Hong Kong Research Grants Council through the General Research Fund (Project no. 11204919).

\appendices

\section{Theoretical Analyses}
\label{appendix_analyses}
\subsection{Proof of Theorem 1}
\label{appendix_theorem_1}

Suppose $F$ is a non-linear transformation in a model. We investigate the effects of layer normalization by analyzing two different implementations:
\begin{align}
    & \mathbf{y}_{l+1}^{(a)} = F_a(\mathbf{y}_l) \label{eq-g1} \\
    & \mathbf{y}_{l+1}^{(b)} = F_b(LN(\mathbf{y}_l)) \label{eq-g2}
\end{align}
where $LN(\cdot)$ is layer normalization. These two implementations are both based on $\mathbf{y}_l$, which is the output from the previous layer. The difference is that Eq.~\ref{eq-g1} does not use layer normalization in the input, while Eq.~\ref{eq-g1} uses it. The gradients of output to input are:
\begin{align}
    & \frac{d \mathbf{y}_{l+1}^{(a)}}{d \mathbf{y}_l} = F_a^\prime (\mathbf{y}_l) \label{eq-de-g1} \\
    & \frac{d \mathbf{y}_{l+1}^{(b)}}{d \mathbf{y}_l}  = F_b^\prime (LN(\mathbf{y}_l)) \cdot LN^\prime(\mathbf{y}_l) \label{eq-de-g2}
\end{align}

There are two differences in Eq.~\ref{eq-de-g1} and Eq.~\ref{eq-de-g2}. $\mathbf{y}_{l}$ is normalized when calculating $F_b^\prime(\cdot)$. It can prevent input from lying in an interval whose gradients are extremely small. Another difference is the additional term $LN^\prime(y_{l})$ in Eq.~\ref{eq-de-g2}. To simplify the notations, we use $\mathtt{x}$ to represent $\mathbf{y}_l$ . Layer normalization is related to the mean and standard deviation among all the dimensions of $\mathtt{x}$. We start the analyses from the $i$-th dimension in $\mathtt{x}$, which is denoted as $\mathtt{x_{[i]}}$.
\begin{equation}
  \begin{aligned}
    LN^\prime(\mathtt{x_{[i]}}) &= [\frac{\mathtt{x_{[i]}} - \mu_\mathtt{x}}{\sigma_\mathtt{x}}]^\prime \\
    &=\frac{([\mathtt{x_{[i]}}-\mu_\mathtt{x}]^\prime \cdot \sigma_\mathtt{x}) - ((\mathtt{x_{[i]}}-\mu_\mathtt{x}) \cdot \sigma_\mathtt{x}^\prime)}{\sigma_\mathtt{x}^2} \label{eq-de-ln}
  \end{aligned}
\end{equation}

The derivative of $[\mathtt{x_{[i]}}-\mu_\mathtt{x}]$ is:
\begin{align}
    [\mathtt{x_{[i]}}-\mu_\mathtt{x}]^\prime &= d(\mathtt{x_{[i]}}-\frac{1}{H}\sum_{j=1}^H \mathtt{x_{[j]}}) / d\mathtt{x_{[i]}} =1-\frac{1}{H}  \label{eq-de-mu}
\end{align}
where $H$ is the dimension of $\mathtt{x}$. The derivative of $\sigma_\mathtt{x}$ is:
\begin{equation}
  \begin{aligned}
 \sigma_\mathtt{x}^\prime &=d\mathbb{E}(\mathtt{x}^2-\mathbb{E}^2(x))^\frac{1}{2} / d \mathtt{x_{[i]}}  \\
 &=\frac{1}{2} \mathbb{E}[\mathtt{x}^2-\mathbb{E}^2(x)]^{-\frac{1}{2}} \cdot
  [\mathbb{E}(\mathtt{x}^2) - \mathbb{E}^2(\mathtt{x})]^\prime \\
 &= \frac{1}{2\sigma_\mathtt{x}}(\frac{2\mathtt{x_{[i]}}}{H} - \frac{2\mathbb{E}(\mathtt{x})}{H}) \\
 &=\frac{\mathtt{x_{[i]}}-\mu_\mathtt{x}}{H \cdot \sigma_\mathtt{x}} \label{eq-de-sigma}
  \end{aligned}
\end{equation}

Considering Eq.~\ref{eq-de-ln},~\ref{eq-de-mu} and~\ref{eq-de-sigma}, we have:
\begin{equation}
  \begin{aligned}
    LN^\prime(\mathtt{x_{[i]}}) &= \frac{ (1-\frac{1}{H}) \cdot \sigma_\mathtt{x} - \frac{(\mathtt{x_{[i]}}-\mu_\mathtt{x})^2}{H \cdot \sigma_\mathtt{x}}}{\sigma_\mathtt{x}^2} \\
    &=\frac{-\frac{1}{H}\sigma_\mathtt{x}^2 + \sigma_\mathtt{x}^2 - \frac{(\mathtt{x_{[i]}}-\mu_\mathtt{x})^2}{H}}{\sigma_\mathtt{x}^3} \\
    &=\frac{-\frac{\sigma_\mathtt{x}^2}{H} + \frac{\sum_{j\neq i}(\mathtt{x_{[j]}}-\mu_\mathtt{x})^2}{H}}{\sigma_\mathtt{x}^3} \\
    &=\frac{-\frac{\sigma_\mathtt{x}^2}{H} + \frac{H-1}{H} \frac{\sum_{j\neq i}(\mathtt{x_{[j]}}-\mu_\mathtt{x})^2}{H-1}}{\sigma_\mathtt{x}^3}
    \label{eq-de-ln-2}
  \end{aligned}
\end{equation}

When $H$ is large enough, we can adopt \textit{the law of large numbers} to obtain the following relation:
\begin{equation}
  \begin{aligned}
    \frac{1}{H-1}{\sum_{j\neq i}(\mathtt{x_{[j]}}-\mu_\mathtt{x})^2} \approx \sigma_\mathtt{x}^2
    \label{eq-de-ln-3}
  \end{aligned}
\end{equation}

Thus, Eq.~\ref{eq-de-ln-2} can be further transformed as:
\begin{equation}
  \begin{aligned}
    LN^\prime(\mathtt{x_{[i]}}) &\approx \frac{-\frac{\sigma_\mathtt{x}^2}{H} + \frac{H-1}{H} \sigma_\mathtt{x}^2}{\sigma_\mathtt{x}^3}  \\
    &=\frac{H-2}{H} \cdot \frac{1}{\sigma_\mathtt{x}}  
    =(1 - \frac{2}{H}) \frac{1}{\sigma_\mathtt{x}}
    \label{eq-de-ln-4}
  \end{aligned}
\end{equation}

We can regard $1 - \frac{2}{H} \approx 1$ when $H$ is large enough, so we have:
\begin{equation}
  \begin{aligned}
    LN^\prime(\mathtt{x_{[i]}}) &\approx \frac{1}{\sigma_\mathtt{x}}
    \label{eq-de-ln-5}
  \end{aligned}
\end{equation}

If the deviation of $\mathtt{x}$ is smaller than 1, this term will be a scaler factor larger than 1. In this time, it can help augment gradients and relieve gradient vanishment problem.

\subsection{Proof of Corollary 1}
\label{appendix_corollary_1}
LSTM is calculated as follows:
\begin{align}
  \begin{pmatrix}
    \mathbf{f}_t \\
    \mathbf{i}_t \\
    \mathbf{o}_t \\
    \mathbf{\hat{c}}_t
  \end{pmatrix} &=\mathbf{W}_h \mathbf{h}_{t-1} + \mathbf{W}_x \mathtt{x}_t + \mathbf{b} \label{eq-lstm-1} \\
  \mathbf{c}_t &= F_s(\mathbf{f}_t) \circ \mathbf{c}_{t-1} + F_s(\mathbf{i}_t) \circ F_h(\mathbf{\hat{c}}_t) \label{eq-lstm-2} \\
  \mathbf{h}_t &= F_s(\mathbf{o}_t) \circ F_h(\mathbf{c}_t) \label{eq-lstm-3}
\end{align}

To simplify analyses, we discuss the influence of the output gate to gradients using following functions:
\begin{align}
  & \mathbf{y}_{l+1}^{(c)} = F_c(\mathbf{\dot{o}}_l \circ \mathbf{\dot{c}}_l) \label{eq-g3} \\
  & \mathbf{y}_{l+1}^{(d)} = F_d(LN(\mathbf{\dot{o}}_l\circ \mathbf{\dot{c}}_l)) \label{eq-g4}
\end{align}
where $\mathbf{\dot{o}}_l$ can be regarded as the output gate in LSTM, so its values are in $(0, 1)$. And $\mathbf{\dot{c}}_l$ can be regarded as the cell state whose values are in $(-1, 1)$. The gradients of these functions are:
\begin{align}
  & \frac{d y_{l+1}^{(c)}}{d \mathbf{\dot{c}}_l} = F_c^\prime (\mathbf{\dot{o}}_l \circ \mathbf{\dot{c}}_l) \cdot diag(\mathbf{\dot{o}}_l) \label{eq-de-g3} \\
  & \frac{d y_{l+1}^{(d)}}{d \mathbf{\dot{c}}_l} = F_d^\prime (LN(\mathbf{\dot{o}}_l \circ \mathbf{\dot{c}}_l)) \cdot LN^\prime(\mathbf{\dot{o}}_l \circ \mathbf{\dot{c}}_l) \cdot diag(\mathbf{\dot{o}}_l)  \label{eq-de-g4}
\end{align}
where $diag(\cdot)$ represents a diagonal matrix.

Both Eq.~\ref{eq-de-g3} and Eq.~\ref{eq-de-g4} contain $\mathbf{\dot{o}}_l$, which is in $(0, 1)$. The gradients related to the calculation of $\mathbf{\dot{c}}_l$ are all scaled down. However, in Eq.~\ref{eq-de-g4}, we have an additional term $LN^\prime(\mathbf{\dot{o}}_l \cdot \mathbf{\dot{c}}_l)$. And the variance of $\mathbf{\dot{o}}_l \cdot \mathbf{\dot{c}}_l$ is:
\begin{equation}
  \begin{aligned}
    Var(\mathbf{\dot{o}}_l \circ \mathbf{\dot{c}}_l) = \mathbb{E}[(\mathbf{\dot{o}}_l \circ \mathbf{\dot{c}}_l)^2] - \mathbb{E}^2(\mathbf{\dot{o}}_l \circ \mathbf{\dot{c}}_l)
  \end{aligned}
\end{equation}

Considering $\mathbf{\dot{o}}_l \in (0,1)$ and $\mathbf{\dot{c}}_l \in (-1,1)$, we have $\mathbb{E}[(\mathbf{\dot{o}}_l \circ \mathbf{\dot{c}}_l)^2]<1$. $\mathbb{E}^2(\mathbf{\dot{o}}_l \circ \mathbf{\dot{c}}_l)$ is a non-negative term, which means $\mathbb{E}^2(\mathbf{\dot{o}}_l \circ \mathbf{\dot{c}}_l) \ge 0$, so we have $Var(\mathbf{\dot{o}}_l \circ \mathbf{\dot{c}}_l) < 1$ and its deviation is also smaller than 1. According to Theorem 1, $LN^\prime(\mathbf{\dot{o}}_l \circ \mathbf{\dot{c}}_l)$ is now an augmentation factor which can offset the influence from $\mathbf{\dot{o}}_l$. This augmentation factor is an adaptive factor. A lower $\mathbf{\dot{o}}_l$ indicates a higher augmentation factor.

\section{Supplementary Explanations about Least Coverage Rate (LCR)}
\label{appendix_lcr}

Least Coverage Rate (LCR) has following features:
\begin{itemize}
  \item It compares two distributions in a fine-grained level. Given two sets of sentences, LCR computes the similarity of every two sentences in the two sets to make sure whether specific modes are covered or not. Thus, it can be more sensitive to the change of sample quality.
  \item The minimum operation in LCR helps it be sensitive to two common problems in generative models: 1) generating samples out of the real distribution; 2) generating samples in high similarities. The coverage rates on test sets are aware of the mode collapse problem, while the coverage rates on inference sets can identify the generated samples out of the real distribution.
  \item It better makes of sentence encoders. Most of sentence encoders are designed to compare sentence similarities with a pre-defined method (e.g., cosine similarity). FED only considers the mean and covariance of two sets, and does not make use of this feature directly.
  \item It is efficient in computation. Although it needs to compute the similarity of every two sentences in the two sets, it can be implemented in a high efficiency way. For example, if we use Universal Sentence Encoder to transform sentences into embeddings, we can easily implement the calculation by making use of matrix multiplication. 
\end{itemize}

\begin{figure*}[htbp]
  \centerline{\includegraphics[scale=0.26]{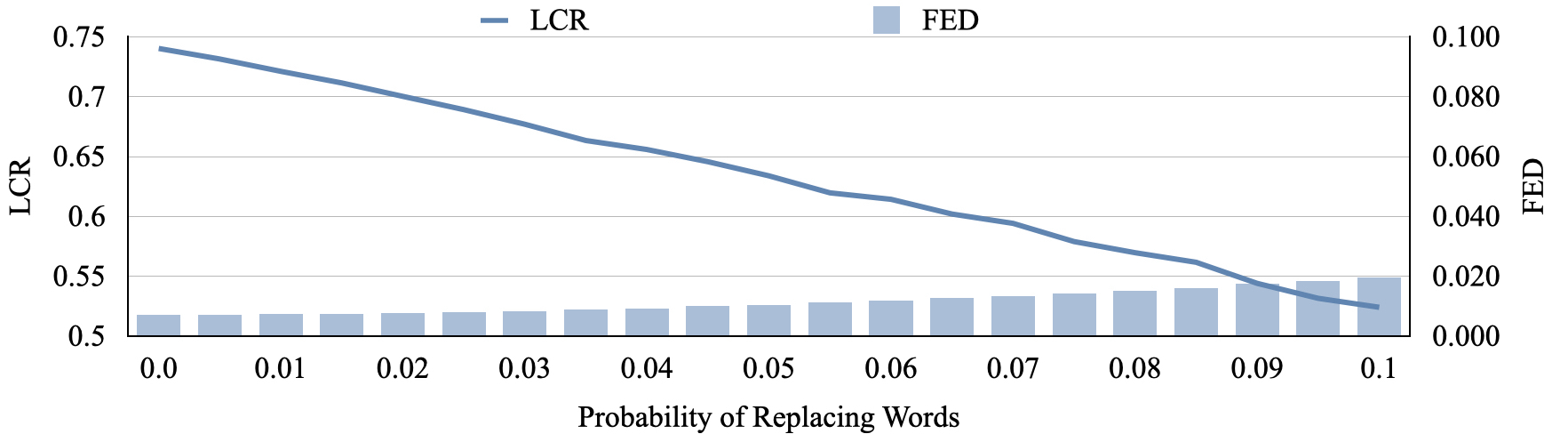}}
  \caption{Changes of LCR and FED on Image COCO Caption Dataset.}
  \label{fig:lcr-fed}
\end{figure*}

\begin{figure}[htbp]
  \centerline{\includegraphics[scale=0.23]{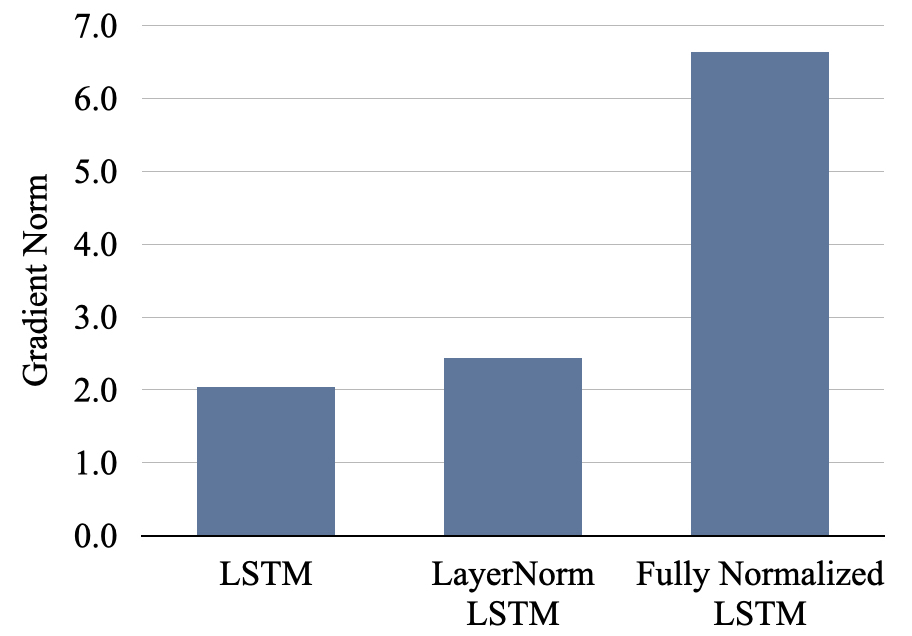}}
  \caption{Gradient Norm.}
  \label{fig:additional-result-2-1}
\end{figure}

We also conduct experiments to explore the effectiveness of Least Coverage Rate (LCR). We explore the sensitivity of LCR and FED to data changes by replacing words in sentences with random ones in a certain probability. The experimental results are shown in Figure~\ref{fig:lcr-fed}.

Generally, the changes of FED are not obvious even when replacing 10\% of words with random words. In the early stage, we nearly cannot observe any changes in FED. It shows that FED is not sensitive to the change of sample quality. Different with FED, LCR is a much more sensitive evaluation metric. Even minor changes can be reflected in LCR. LCR decreases from around 0.75 to less than 0.5 when 10\% of words are replacing with random words. This result shows that LCR can be a good compliment when the compared model gets close performance in FED.

Least coverage rate uses the minimum value among the coverage rates of test sets and inference sets as result. In practice, we can give more detailed analyses about these two coverage rates to reveal the problems hidden in models. When the coverage rate of the inference set is higher, it shows the model tends to generate samples out of real distributions. If the coverage rate of the test set is higher, it demonstrates the mode collapse problem in the model.

\section{Experiment Details}
\label{appendix-exp_details}

The aligner is constructed based on the original Transformer structure. The representations from the aligner or the generator will be added with fixed positional encoding and then feed into a stack of Transformer blocks. Each block is consisted of a multi-head attention layer and feed forward network layer. Layer normalization is adopted after each layer. Both the generator and the discriminator are constructed based on the fully normalized LSTM.

The hyperparameters of InitialGAN are:
\begin{itemize}
  \item Bate size: 128
  \item Maximum epoch of the aligner training: 200
  \item Maximum epoch of the adversarial training: 3,000 for COCO, 2,000 for EMNLP
  \item Embedding size: 512
  \item Feature size in the aligner: 512
  \item Feature size in the discriminator: 1024
  \item Feature size in the generator: 512
  \item Dimension of random noise: 128
  \item Layer number of the aligner: 4
  \item Layer number of the discriminator: 2
  \item Layer number of the generator: 2
  \item Head number of the aligner: 8
  \item Dropout rate for the aligner: 0.5
  \item Dropout rate in dropout sampling: 0.75
  \item Learning rate of the aligner: 0.0001
  \item Learning rate of the discriminator: 0.0004 for COCO, 0.0002 for EMNLP
  \item Learning rate of the generator: 0.0001
  \item $\lambda_a$ in the aligner loss: 0.1
  \item $\lambda_d$ in the discriminator loss: 50
  \item Optimizer of the aligner: AdamW ($\beta_1=0.9$, $\beta_2=0.999$, weight decay=0.00001)
  \item Optimizer of the discriminator: AdamW ($\beta_1=0.5$, $\beta_2=0.9$, weight decay=0.0001)
  \item Optimizer of the generator: Adam ($\beta_1=0.5$, $\beta_2=0.9$)
\end{itemize}

\begin{table}[]
  \centering
  \caption{Time and Spatial Complexity of LayerNorm LSTM and Fully Normalized LSTM}
  \label{table:time_spatial}
  \begin{tabular}{|c|c|c|}
  \hline
  \textbf{Model}                 & \textbf{Parameter number} & \textbf{Computation time} \\ \hline
  LayerNorm LSTM        &  2,108,416     &   109.4ms     \\ \hline
  Fully Normalized LSTM &  2,109,440     &   138.5ms      \\ \hline
  \end{tabular}
\end{table}

\begin{figure}[htbp]
  \centerline{\includegraphics[scale=0.23]{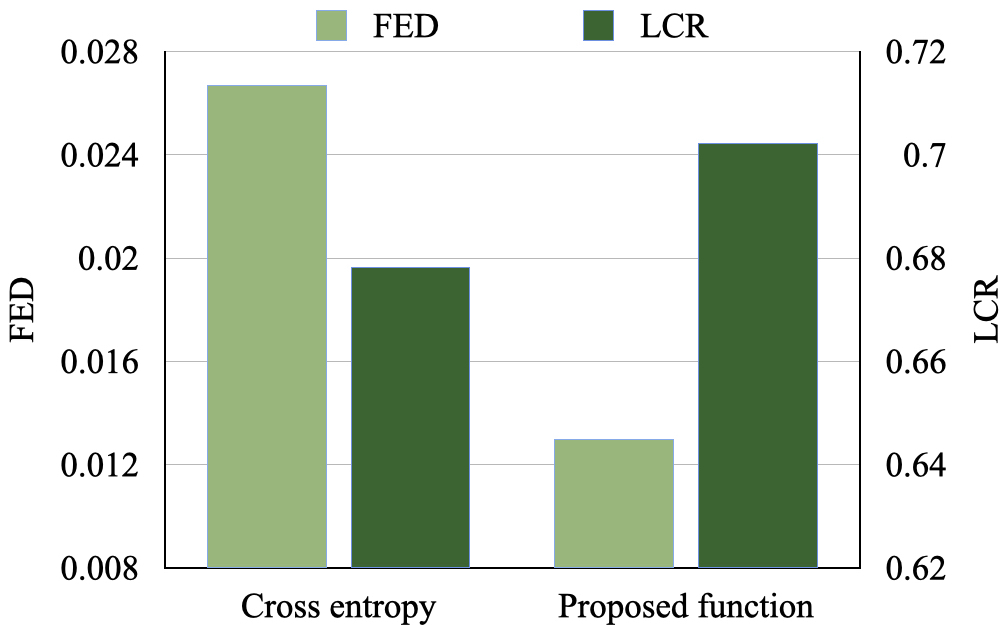}}
  \caption{The performance of InitialGAN with aligners in different objectives.}
  \label{fig:additional-result-2-2}
\end{figure}

We save the model after every epoch, and select the best model based on FED on validation sets. We follow the settings of the previous work~\cite{DBLP:conf/sigir/ZhuLZGZWY18} to implement MLE except for the layer number and feature size which are set to be same with the generator in InitialGAN. It not only leads the model to be more comparable, but also improves its performance. We obtain the results of other language GANs by running the public code~\footnote{\url{https://github.com/geek-ai/Texygen}}~\footnote{\url{https://github.com/deepmind/deepmind-research/tree/master/scratchgan}}~\footnote{\url{https://github.com/weilinie/RelGAN}}. We implement InitialGAN based on Tensorflow 2~\cite{tensorflow2015-whitepaper}. It is trained on NVIDIA GeForce RTX 3090 with around 12G RAM.

\section{Supplementary Experimental Results}
\label{appendix-exp_results}

According to our analyses, fully normalized LSTM can relieve the gradient vanishment by providing an augmentation term. We train models with different variants of LSTM, and show their average gradient norms of first 100 training batches in Figure~\ref{fig:additional-result-2-1}. These norms are calculated based on the gradients of the input linear transformation matrix in the last layers of generators. LayerNorm LSTM can slightly augment the gradient norm, while fully normalized LSTM can obtain more obvious augmentation. The experimental results are consistent with our theoretical analyses. 

Besides, we also conduct experiments to compare the time and spatial complexity of LayerNorm LSTM and fully normalized LSTM. We construct 1-layer model with these two structures and the hidden size is set to be 512. We feed a sequence of length 50 into the models 1,000 times and calculate the average running time. The results are shown in Table~\ref{table:time_spatial}. The parameter numbers of Fully Normalized LSTM and LayerNorm LSTM are extremely close. The computation time of Fully Normalized LSTM is slightly higher than LayerNorm LSTM because of the additional layer normalization operation.

\begin{figure}[htbp]
  \centerline{\includegraphics[scale=0.22]{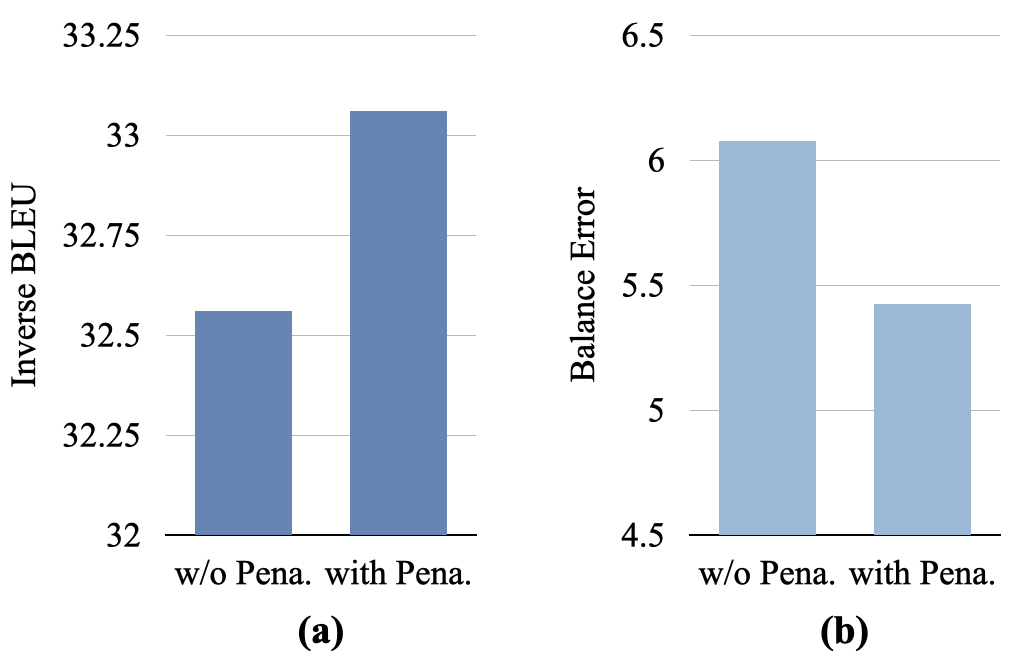}}
  \caption{Effectiveness of the penalization term (Pena.) in the training objective of the aligner. (a) Inverse BLEU. (b) Balance error. Lower is better.}
  \label{fig:additional-result-3-1}
\end{figure}

\begin{figure}[htbp]
  \centerline{\includegraphics[scale=0.22]{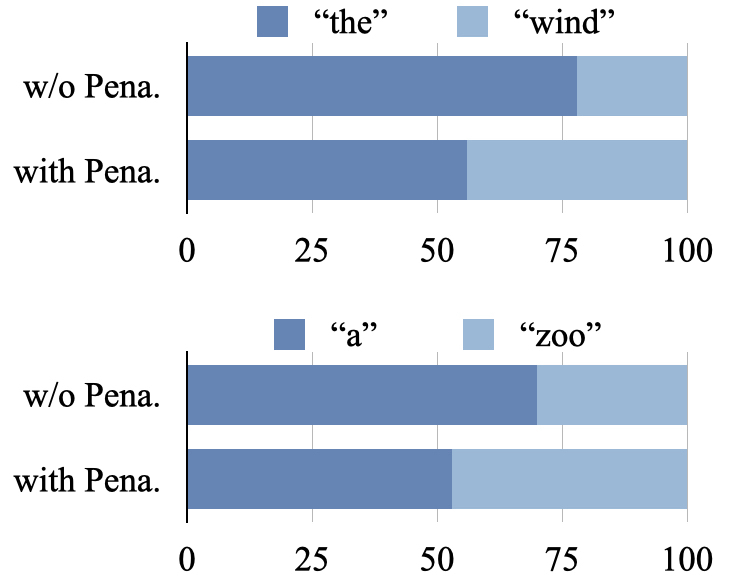}}
  \caption{Mapping times of two pairs of words.}
  \label{fig:additional-result-3-2}
\end{figure}

In the aligner, we map words into distributions instead of specific embeddings. We show the FED and LCR ($\tau=0.65$) of these two mapping methods on Image COCO Caption Dataset in Figure~\ref{fig:additional-result-2-2}. Aligner trained by cross entropy, which maps words into specific points, is inferior to the one trained by our proposed loss function in both FED and LCR. Cross entropy does not provide confirmed mapping relations when the generated representations are away from the real representations. Minor errors in the generated representations may lead them to be mapped into totally irrelevant words. It adds noise to training and increases instability in inference.

In the training objective of the aligner, we add a penalization term to constraint words in different frequency having the regions in similar size. We also explore the effectiveness of this penalization term. We firstly compare its influence to Inverse BLEU on Image COCO Caption Dataset. As shown in Figure~\ref{fig:additional-result-3-1} (a), the penalization term helps the model to get better Inverse BLEU. Besides, we also try to evaluate the size of different word regions directly. Thus, we randomly select the representations of two words and use linear interpolation to get 100 points between these two representations. Then, we use the transformation layer to transform these representations back into words. We denote the times that a word is mapped as mapping times. If the size of the two regions are similar, the mapping times of the two words should be generally same, i.e., close to 50, so we design a balance error to evaluate the size of the regions:
\begin{equation*}
  balance\_error = \frac{1}{n} \sum_{i=1}^n|times_i - exp\_times|
\end{equation*}
where $times_i$ is the mapping times of a word in the $i$-th word pair (choosing any of the word in the pair will give same result), and $exp\_times$ is the expected mapping times which is 50 in our experiments. We randomly select 1,000 word pairs to conduct this experiment and the results are shown in Figure~\ref{fig:additional-result-3-1} (b). Penalization term helps the model gets lower balance error. It means the regions of different words are more balanced after penalization. We also show two cases of mapping times in Figure~\ref{fig:additional-result-3-2}. The mapping times of these words are all closer to 50 after penalization. These experiments show the effectiveness of the penalization term.

\section{Unsuccessful Attempts}
\label{appendix-fail_attempts}

We show some attempts that do not bring improvements in our experiments so as to provide a reference for future work:

\begin{itemize}
  \item Using Bi-LSTM to build the discriminator. It does not get improvements.
  \item Using spectral normalization to regularize the discriminator. It is unsuccessful.
  \item Using Hinge loss as the training objective. It is unsuccessful when adopting either spectral normalization or gradient penalty.
  \item In addition to the original loss, adding next word prediction or part-of-speech (POS) prediction to train the discriminator by making use of multi-task learning. It is unsuccessful.
  \item Using dropout in the discriminator. It can not get improvements.
  \item Adding random noise in each layer of the discriminator and the generator. It is unsuccessful.
  \item Using AdamW as the optimizer of the generator. Mode collapse is found.
  \item Prompt the model to firstly learn first few words and adds the number of words gradually. It is unsuccessful.
  \item Learning rates of the discriminator and the generator are decayed during training process. It is unsuccessful.
\end{itemize}

\section{Generated Samples}
\label{appendix-samples}
We demonstrate generated samples given by the models trained on COCO Image Caption Dataset in Table~\ref{table:coco_sample}.

\begin{table*}[]
  \centering
  \caption{Samples Generated by Models Trained on COCO Image Caption Dataset}
  \label{table:coco_sample}
  \begin{tabular}{p{16cm}}
  \hline
  \multicolumn{1}{c}{\textbf{MLE}} \\ \hline
  - a living room has no parking watch a chair . \\
  - a man is playing wii in a home hand . \\ 
  - two groups of skiers are hanging around the large business park . \\ 
  - bikes outside by a big bridge .\\ 
  - a white plate topped with lots of chinese food .   \\ \hline

  \multicolumn{1}{c}{\textbf{SeqGAN}} \\ \hline
  - there is a basket in the small bathroom with a white sink , and bathroom .  \\ 
  - the side of a small dog is sitting next to it . \\
  - there is sitting in his hand up in the water .  \\
  - the side of a pole has a building is in the rear view mirror . \\ 
  - three donuts on the plate of broccoli are in the green .                  \\ \hline

  \multicolumn{1}{c}{\textbf{RankGAN}} \\ \hline
  - a black and white photo of a toiletry shelf in a room .    \\
  - a black cat watches a city from the corner of a room .    \\
  - a street with a white and yellow motorcycle next to a street .   \\
  - the zebra stands on a grass covered mountain .    \\
  - a cart holding a note on a ground .    \\ \hline

  \multicolumn{1}{c}{\textbf{MaliGAN}} \\ \hline
  - a row of traffic lights showing off the side of an intersection .   \\ 
  - a man walks behind them in high chair with a wii remote in her hands in a nursery room . \\ 
  - a row of pink fire trucks parked outside by a row in the corner of the head of a window .    \\ 
  - a row up on a bus driving over a city park pair of scissors in a donut .    \\  
  - a city is floating in the sky .      \\ \hline

  \multicolumn{1}{c}{\textbf{LeakGAN}} \\ \hline
  - a man in a small corner on a counter and a table . balcony .   \\ 
  - a man and a baseball and a photograph of wine . word throw a wave .  \\ 
  - a man is on a busy city street with a pink . choices on the beach . \& sam .  \\
  - a man riding a horse track with lots of items on a wooden desk . slot . \\ 
  - a man riding a skateboard while holding a tennis racket . describing is on it . lined up . dancer      \\ \hline

  \multicolumn{1}{c}{\textbf{ScratchGAN}} \\ \hline
  - people and standing in a large yard playing frisbee . \\                                                              - a device with smaller sandwiches rusted between one hair .       \\                                                                   - a person sitting on on the park looking at two pairs seductive .     \\ 
  - a grasses sticking his head over a train on the tracks .   \\ 
  - a 2 machine is cutting into the air a ride .     \\ \hline

  \multicolumn{1}{c}{\textbf{RelGAN}} \\ \hline
  - a woman sitting on a bench next to a man .  \\
  - a woman holding up a smart phone in front of a man .  \\
  - two women sitting on the beach with two horses .  \\
  - a man in a black and white cat lays on a corner .  \\
  - a man in an orange and white cat laying on a carpet of a sink .    \\ \hline

  \multicolumn{1}{c}{\textbf{InitialGAN}} \\ \hline
  - a man walks down the street holding a surfboard .   \\ 
  - a man in a jacket playing with his frisbee in his hand .   \\ 
  - a baby elephant in the grass with a toy zoo .   \\
  - a cute teddy bear is wearing a red shirt .   \\ 
  - a dog playing a game of baseball on a sunny day .   \\ \hline
  \end{tabular}
\end{table*}

\bibliographystyle{IEEEtran}
\bibliography{references}

\vfill

\end{document}